\documentclass[11pt,a4paper]{article}
\usepackage[affil-it]{authblk}
\usepackage[hyperref]{acl2019}
\usepackage{times}
\usepackage{latexsym}
\usepackage{graphicx}
\usepackage{subfigure}
\usepackage{url}

\usepackage{amsmath,amsfonts,bm}

\def\eqref#1{equation~\ref{#1}}

\def\1{\bm{1}}

\def\vtheta{{\bm{\theta}}}

\def\vh{{\bm{h}}}

\def\vt{{\bm{t}}}
\def\vu{{\bm{u}}}

\DeclareMathAlphabet{\mathsfit}{\encodingdefault}{\sfdefault}{m}{sl}
\SetMathAlphabet{\mathsfit}{bold}{\encodingdefault}{\sfdefault}{bx}{n}

\DeclareMathOperator*{\argmax}{arg\,max}

\DeclareMathSymbol{*}{\mathbin}{symbols}{"03}
\DeclareMathSymbol{\ast}{\mathbin}{symbols}{"03}

\usepackage{boxhandler}
\usepackage{float}
\usepackage{booktabs} 
\usepackage{adjustbox}

\usepackage{linguex}

\newcommand{\todo}[1]{}
\renewcommand{\todo}[1]{{\color{red} {#1}}}
\newcommand{\aclnew}[1]{}
\renewcommand{\aclnew}[1]{{\color{blue} {#1}}}
\usepackage{amssymb}
\usepackage{amsmath}

\aclfinalcopy 

\title{Word-order biases in deep-agent emergent communication}

\author[1,2]{Rahma Chaabouni}
\author[1]{Eugene Kharitonov}
\author[1]{Alessandro Lazaric}
\author[1,2]{\\Emmanuel Dupoux}
\author[1,3]{Marco Baroni}
\affil[1]{Facebook A.I. Research}
\affil[2]{Cognitive Machine Learning (ENS - EHESS - PSL Research University - CNRS - INRIA)}
\affil[3]{ICREA}
\affil[ ]{\tt {\{rchaabouni,kharitonov,lazaric,dpx,mbaroni\}@fb.com}}

\date{}

\begin{document}
\maketitle

\newcommand\dpx[1]{{\color{blue}{ED: #1}}}

\begin{abstract}

 
Sequence-processing neural networks led to remarkable progress on many NLP tasks. As a consequence, there has been increasing interest in understanding to what extent they process language as humans do. We aim here to uncover which biases such models display with respect to ``natural" word-order constraints. We train models to communicate about paths in a simple gridworld, using miniature languages that reflect or violate various natural language trends, such as the tendency to avoid redundancy or to minimize long-distance dependencies. We study how the controlled characteristics of our miniature languages affect individual learning and their stability across multiple network generations. The results draw a mixed picture. On the one hand, neural networks show a strong tendency to avoid long-distance dependencies. On the other hand, there is no clear preference for the efficient, non-redundant  encoding of information that is widely attested in natural language. We thus suggest inoculating a notion of ``effort'' into neural networks, as a possible way to make their linguistic behaviour more human-like.
 
\end{abstract}

\section{Introduction}
\label{sec:introduction}

Deep neural networks, and in particular ``sequence-to-sequence''  \cite[Seq2Seq,][]{Sutskever:etal:2014} LSTM
recurrent networks, attained astounding successes in many linguistic
domains \cite{Goldberg:2017}, but we still have a poor understanding of
their language processing mechanisms \cite{Lake:Baroni:2017}. 
We study here whether word-order constraints commonly observed in natural
language are also found as ``inductive" biases in
recurrent networks. We consider three such constraints. The first is
temporal \emph{iconicity}, defined as the tendency of clauses denoting
events to reflect the chronological order of the denoted events
\citep[as in Caesar's \emph{veni, vidi,
  vici};][]{Greenberg:1963,Haiman:1980,Newmeyer:1992,Radden:Dirven:2007,Diessel:2008,Marcus:Calude:2010,deRuiter:etal:2018}. The
second is the need to disambiguate the role of sentence constituents,
that can be achieved either by means of fixed-word order (e.g., in an
SVO language the first noun phrase denotes the subject), or by overting
morphological markers (e.g., the subject is marked with nominative
case). As the two mechanisms are redundant, a trade-off is generally
observed, where languages preferentially adopt one or the other
\citep{Comrie:1981,Blake:2001}. Finally, we consider the general
tendency of languages to avoid or minimize long-distance dependencies
\cite{Hawkins:1994,Gibson:1998,Futrell:etal:2015}. As \newcite{Futrell:etal:2015} observe, ``I
checked [it] out'', with one word intervening between the verb and the
particle it composes with, \textit{`is easier or
more efficient to produce and comprehend'} than ``I checked [the place
you recommended] out'', with four intervening words.

We test whether such constraints affect LSTM-based Seq2Seq models. To this end, we train them as agents in a simple $2D$ gridworld environment, in which they give and receive navigation instructions in hand-designed artificial languages satisfying or violating the constraints.
We first study which languages are harder to learn for individual agents. Then, we look at the cultural transmission of language characteristics through multiple agent generations by means of the iterated learning paradigm \cite[]{Kirby:etal:2014}.\footnote{Code link: \url{https://github.com/facebookresearch/brica}.}

Our results suggest a mixed picture. LSTM agents are partially
affected by natural constraints, both in terms of learning difficulty
and stability of patterns through evolution. For example, they show a strong tendency to avoid long-distance dependencies. Still, some patterns are considerably
different from those encountered in human language. In particular,
LSTMs generally have a preference for the reverse version of an iconic
language, and only show a weak tendency towards avoidance of redundant
coding.

\section{Related work}
\label{sec:related}

There is increasing interest in applying methods from linguistics and
psychology to gain insights on the functioning of language processing
networks, as witnessed by the recent BlackBoxNLP workshop at EMNLP 2018
\cite{Linzen:etal:2018}. In this context, researchers have looked at how \emph{trained} models solve different NLP tasks characterizing their outputs and internal representation. We instead focus directly on uncovering their ``innate'' biases \emph{while learning a task}.

We study whether LSTM-based Seq2Seq models deployed as communicating
agents are subject to some of the natural pressures that characterize
the typology and evolution of human languages. In this respect, we
connect to the recent research line on language emergence in deep
network agents that communicate to accomplish a task
\cite[e.g.,][]{Jorge:etal:2016,Havrylov:Titov:2017,Kottur:etal:2017,Lazaridou:etal:2017,Choi:etal:2018,Evtimova:etal:2018,Lazaridou:etal:2018,mordatch2018}. Most
of this work provides the agents with a basic communication channel,
and evaluates task success and the emerging communication protocol in
an entirely bottom-up fashion. We train instead our agents to
communicate with simple languages possessing the properties we want to
study, and look at whether such properties make the languages easier
or harder to learn. Other studies \cite{Lee:etal:2018,lee2017} had also seeded their agents
with (real) languages, but for different purposes (letting them
develop translation skills).

We introduce miniature artificial languages that respect or violate specific constraints. Other studies have used such languages with human subjects to test hypotheses about the origin of cross-linguistically frequent patterns \cite[see][for a survey]{Fedzechkina2016}. We follow this approach to detect biases in \emph{Seq2Seq models}. We specifically rely on two different measures. First, we evaluate the speed of learning a particular language, assuming that the faster it is, the easier its properties are for the agent \cite[e.g.,][]{Tily2011, Hupp2009}. Second, we look at the cultural evolution of a language by means of the \emph{iterated language learning} paradigm \cite[see][for a survey]{Kirby:etal:2014}. That is, we investigate the changes that modern Seq2Seq networks exposed to a language through multiple generations introduce, checking which biases they expose.

\section{Experimental setup}
\label{sec:setup}

\subsection{Languages}
\label{sec:languages}

Our environment is characterized by trajectories of 4 oriented actions
(LEFT, RIGHT, UP, DOWN). A trajectory contains from 1 to 5 segments,
 each composed of maximally  3 steps in the same
direction. A possible 3-segment trajectory is: LEFT LEFT
RIGHT UP UP UP, with (LEFT LEFT), (RIGHT), and (UP UP UP) being its segments.

\paragraph{Fixed- and free-order languages} In a \textit{fixed-order}
language, a segment is denoted by a phrase made of a command (C) and
a quantifier (Q). An utterance specifies an order for the
phrases. For example, in the \emph{forward-iconic} language,
3-phrase utterances are generated by the following rules:


\ex. U $\to$ P1 P2 P3\\
     P(1$|$2$|$3) $\to$ C Q\\
     C $\to$ (left$|$right$|$up$|$down)\\
     Q $\to$ (1$|$2$|$3)\label{ex:iconic}

Shorter and longer utterances are generated analogously
(a N-phrase utterance always has form P1 P2 $\ldots$
PN). Importantly, the interpretation function associates PN
to the N-th segment in a trajectory, hence the temporal iconicity
of the grammar. For example, the utterance ``left 2 right 1 up
3'' denotes the 3-segment trajectory: LEFT LEFT RIGHT UP UP UP.

The \emph{backward-iconic} language is analogous, but phrases are
interpreted right-to-left. \emph{Non-iconic} languages use the same
interpretation function associating PN to the N-th segment, but now
the grammar licenses phrases in a fixed order different from that of
the trajectory. For example, 3-phrase utterances might be generated by
U $\to$ P2 P3 P1 (the trajectory above would be expressed by: ``right
1 up 3 left 2''). Relative phrase ordering  is fixed across
utterances irrespective of length. For example, 2-phrase utterances in
the language we just illustrated must be generated by U$\to$P2 P1, to
respect the fixed-relative-ordering constraint for P2 and P1 with
respect to the 3-phrase rule.

Fixed-order languages \textit{with (temporal ordering) markers} use
the same utterance rules, but now each phrase PN is also associated
with an unambiguous marker. For example, the \emph{iconic+markers}
language obeys the first rule in \ref{ex:iconic}, but the phrases are
expanded by:

\ex. P1 $\to$ first C Q\\
     P2 $\to$ second C Q\\
     P3 $\to$ third C Q\label{ex:iconic_markers}

In the iconic+markers language, the trajectory above is expressed by ``first left 2 second right 1 third up 3''.

A \textit{free-order} language licenses the same phrase structures as
a fixed-order language and it uses the same interpretation function,
but now there are rules expanding utterances with all possible phrase
permutations (e.g., 3-phrase utterances are licensed by 6 rules: U
$\to$ P1 P2 P3, U $\to$ P1 P3 P2, $\ldots$).\footnote{Equivalently, a
  free-order language is generated in two stages from a fixed-order
  one through a scrambling process.}  Both ``second right 1 third
up 3 first left 2'' and ``third up 3 second right 1 first left 2'' are
acceptable utterances in the free-order language with markers. 
Examples of trajectory-to-utterance mappings of these artificial languages are provided in Supplementary

\paragraph{Long-distance language} 
We consider a long-distance
language where any phrase can be split and wrapped around a single
other phrase so that a long-distance dependency is created between the components of the outermost phrase.\footnote{
  Note also that this language is projective, excluding cross-dependencies.} We
treat long-distance dependencies as optional, as in languages in which
they are optionally triggered, e.g., by information structure
factors. We compare the \emph{long-distance} language to a
\emph{local} free-order language lacking the long-distance split
construction. Since the long-distance option causes a combinatorial
explosion of possible orders, we limit trajectories to 3
segments.  At the same time, to have two languages partially
comparable in terms of variety of allowed constructions, we extend the
grammars of both to license free order within a phrase. 
Finally, markers are prefixed to both the
command and the quantifier, to avoid ambiguities in the long-distance
case. Summarizing, the local language is similar to the
free-order+markers one above, but markers are repeated before each
phrase element, and extra rules allow the quantifier to precede or
go after the
command, e.g., both of the following structures are permitted: P1
$\to$ first Q first C; P1 $\to$ first C first Q (``first left first
2''; ``first 2 first left''). The long-distance grammar further
includes rules where P1 has been split in two parts, such as:

\ex. U $\to$ first C1 P2 first Q1 P3\\
     U $\to$ first Q1 P2 first C1 P3\label{ex:long-distance}

with C1 and Q1 expandable into the usual terminals (LEFT,
RIGHT\ldots and 1, 2, 3, respectively).\footnote{Equivalently, long-distance constructions are derived by movement rules from canonical underlying structures.} The interpretation function associates a discontinuous $\{$CN, QN$\}$  phrase
 with the N-th segment in the
trajectory. The first rule in \ref{ex:long-distance}
licenses the utterance ``first left second right second 1
first 2 third up third 3'', denoting the example trajectory at
the beginning of this section. Similar rules are introduced for all possible splits
of a phrase around another phrase (e.g., the elements of P2
around P1, those of P1 around P3, etc.). Only one split is
allowed per-utterance. Examples of trajectory-to-utterance mappings in the long and local-distance languages are provided in Supplementary.

\paragraph{Datasets} We generate sentences associated to all possible
trajectories in the environment ($88572$ in the fixed- and free-order
language environment, $972$ in the local- and long-distance
environment experiments).  We randomly split all possible distinct
trajectory-utterance pairs into training ($80\%$) and test/validation
sections ($10\%$ each).

\setcounter{equation}{3}

\subsection{Models}
\label{sec:Archi}

\paragraph{Architecture} The agents are Encoder-Decoder Seq2Seq
architectures~\citep{Cho2014,Sutskever:etal:2014} with single-layer
LSTM recurrent units~\citep{Hochreiter:Schmidhuber:1997}. In light of
the interactive nature of language, an agent is always trained to be
both a \emph{Speaker}, taking a trajectory as input and producing an
utterance describing it, and as a \emph{Listener}, executing the
trajectory corresponding to an input utterance. Input and output
vocabularies are identical, and contain all possible actions and
words.\footnote{Word and action symbols are disjoint, e.g., the action symbol `LEFT' is different from the word symbol
  'left'.} %
When an agent plays the Speaker role, it uses input action representations and output word representations, and conversely in
the Listener role. 
We tie the embeddings of the encoder input and of the decoder output
\cite{Press2016} making input and output representations of
words and actions coincide. As a result, Speaker training affects the
representations used in Listener mode and \emph{vice versa}.
Experiments without tying (not reported) show similar results with slower convergence.
We additionally explore a standard attention mechanism
~\citep{Bahdanau2014}.

\paragraph{Training}
\label{sec:Training}
We consider two scenarios. In \emph{individual learning}, an agent is
taught a  language by interacting
with a hard-coded ground-truth ``teacher'', represented by the
training corpus.  In the \emph{iterated learning} setup, a lineage of
agents is trained to speak and listen by interacting with a ``parent''
agent. After convergence, an agent is fixed and used as a parent to
train the next child.

\paragraph{Individual learning} 
We synchronously train the agent to speak (from trajectory $\vt$ to
utterance $\vu$) and listen (from utterance $\vu$ to trajectory
$\vt$). Training the Listener is similar to standard
Seq2Seq training with teacher forcing~\cite[p.~376]{Goodfellow2016}.  We change the training
procedure for the Speaker direction, as we must handle one-to-many
trajectory-to-utterance mappings in free-order languages. We 
describe it below.

For each trajectory, we consider all corresponding utterances equally probable.  Given a trajectory input, an agent
must be able to produce, with equal probability, all utterances that correspond
to the input.  To achieve this, taking  inspiration from the
multi-label learning literature, we fit the agent's output
distribution to minimize KL-divergence from the
uniform over target utterances. We adopt the ``Na\"ive'' method
proposed by ~\newcite{Jin2003} (see Supplementary for how we derive
the loss function in Eq.~(\ref{eq:pre-train})).

Formally, our languages map trajectories $\vt_j$ to one (fixed-order) or
multiple (free-order) utterances $\{\vu\}_j = \{\vu_j^1, \vu_j^2, \ldots\}$. 
The trajectory $\vt$ is fed into the
encoder, which produces a  representation of the action
sequence. Next, the latter is fed into the decoder along with
the start-of-the-sequence element $u_0 = sos$. At each step, the
decoder's output layer defines a categorical distribution
$p_{\vtheta}(u_k | u_{k-1}, \vh_{k})$ over the next output word
$u_k$. This distribution is conditioned by the previous word $u_{k-1}$
and the hidden state $\vh_k$. 
As with the Listener, we use teacher forcing, so that the distribution of each
word is conditioned by the ground-truth terms coming before it.


Overall, the model parameters $\vtheta$ are optimized to minimize the loss $\mathcal{L}$ over $(\vt_j, \{\vu\}_j)$:
\setcounter{equation}{3}
\begin{equation}
\label{eq:pre-train}
    \mathcal{L} =  -\sum_j \frac{1}{n_j}\sum_{\vu \in \{\vu\}_j} \sum_{k=1}^{|\vu|} \log p_{\vtheta}(u_{k} | u_{k-1}, \vh_{j, k})
\end{equation}
In Eq.~(\ref{eq:pre-train}), $n_j$ denotes the number of target
utterances for the $j$th example, $n_j = |\{\vu\}_j|$; $\vu$ iterates
over the utterances $ \{\vu\}_j$; and $u_{k}$ enumerates words in the
utterance $\vu$ as $k$ varies. As the number of ground-truth
utterances $\{\vu\}_j$ can be high, we sub-sample $n=6$ when training
free- and fixed-order languages.\footnote{Sampling is trivial in the
  latter case, since $\{\vu\}_j$ contains a single utterance. Note
  that in this case the loss $\mathcal{L}$ reduces to the
  negative log-likelihood. This allows us to use the same loss
  function for free- and fixed-order languages.} This
considerably speeds up training without significantly harming
performance. We use all the possible utterances when training on
long-distance languages ($n$ equals the the number of all possible utterances).


For all studied languages, we perform a grid search over hidden layer
[$16$,$20$] and batch sizes [$16$,$32$], and report test set results of
the best validation configuration for each language re-initialized
with $5$ different seeds.  We stop training if development set accuracy does not
increase for 5 epochs or when 500 epochs are reached. In all scenarios, the optimization is performed with the
\textit{Amsgrad}~\cite{j.2018on}  which is an improved version of the standard Adam~\cite{Kingma2014}; we did not experiment with other optimizers.
We use the algorithm with its default parameters, as implemented in Pytorch~\cite{pytorch}.

\begin{figure}[tb]
\centering
{
    \includegraphics[width=\columnwidth, keepaspectratio]{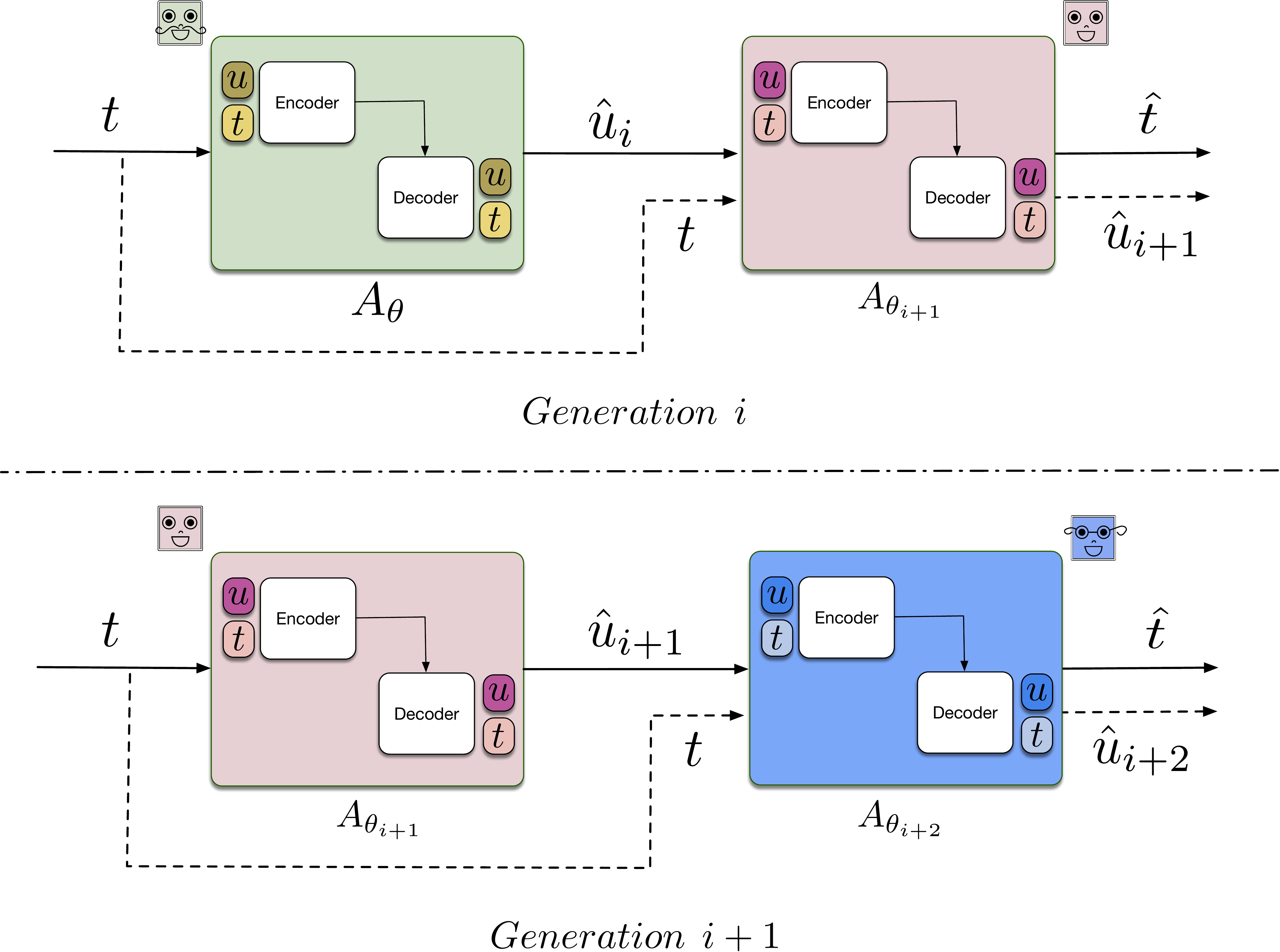}
}
\vspace{-1.2\baselineskip}
\caption{\textbf{Iterated learning}. Language is transmitted to a child agent
  $A_{\vtheta_{i+1}}$ by teaching it to \emph{speak} imitating the
  utterances of parent $A_{\vtheta_i}$ given the same input
  trajectories (dashed lines) and to \emph{listen} to the parent
  utterances, converting them to trajectories (continuous
  lines). After training, former child $A_{\vtheta_{i+1}}$ becomes
  the parent of a new agent
  $A_{\vtheta_{i+2}}$. \vspace{-1.4\baselineskip}}\label{fig:iterated_learning}
\end{figure}

\paragraph{Iterated learning} At ``generation 0'' agent
$A_{\vtheta_0}$ is trained individually as described above. Once $A_{\vtheta_0}$ is trained, we fix its
parameters and use it to train the next-generation agent,
$A_{\vtheta_1}$. $A_{\vtheta_1}$, after training, is in its turn fixed and used to
train the next agent $A_{\vtheta_2}$, etc.  At each iteration, the
child agent $A_{\vtheta_{i+1}}$ is trained to \emph{imitate} its
parent $A_{\vtheta_{i}}$ as follows.  Suppose that, given $\vt$, the
parent agent produces $n$\footnote{We use the same number $n$ defined in individual learning section.} utterances $\{\hat \vu\}=\{\hat \vu^1, \hat \vu^2, ...\hat \vu^n\}$ (these utterances are obtained by sampling from the parent's decoder and can be identical). 
Then, we train the child agent to:
(a) \emph{listen}: map each utterance $\hat \vu^j$ to the trajectory
$\vt$, and (b) \emph{speak}: given the trajectory $\vt$, produce the
utterance $\hat \vu$ that 
is within $\{\hat \vu\}$ (Fig.~\ref{fig:iterated_learning}). Importantly, even if the parent's parameters are fixed at each generation, the child agent is allowed, while achieving perfect accuracy, to introduce changes into its' parent language, making the latter more closely aligned with its ``innate'' biases. \footnote{as exemplified in the experiments below, the child can reach perfect accuracy while having a different distribution over the utterances than its parent.}

Importantly,
the language is not forced to remain stationary across
generations.

\paragraph{Evaluation}

We evaluate agents both as Listeners and as Speakers. The former is
standard, as each input $\vu$ maps to a single output $\vt$. Since the
Speaker can be one-to-many, in order to obtain a single prediction
$\vu$ given trajectory $\vt$, we
predict at each time step $k$ a word $u^{\ast}_k=\argmax_{u_k}(p_{\vtheta}(u_k | u^{\ast}_{k-1}, \vh_{k}))$. 
This word is fed to the next unit of the decoder, and so on
until $u^{\ast}_K = eos$. The final prediction $\hat{\vu}^{\ast}$ is
then defined as the sequence $[u^{\ast}_1, u^{\ast}_2 ...u^{\ast}_K]$, and compared to $M$
\emph{samples} from the true distribution $P(\vu|\vt)$.  If
$\hat{\vu}^{\ast}$ matches \emph{one} of the true samples, the agent
succeeds, otherwise it fails (in iterated learning, $P(\vu|\vt)$
corresponds to the parent's distribution). In other
words, we are not evaluating the model on a perfect fit of the ground-truth (parent's, in case of iterated learning)
distribution, but we score a hit for it as long as it outputs a
combination in $P(\vu|\vt)$. This mismatch between the training and evaluation criteria allows the emergence of interesting patterns (as we allow the agent to drift from the ground-truth distribution) while constituting a reasonable measure of actual communication success (as the agent produces an utterance that is associated to the input trajectory in the ground-truth).

\section{Experiments}
\label{sec:experiments}

\begin{figure*}[ht]
\centering
\vspace{-1.3\baselineskip}
\hspace{-2.5\baselineskip}
\subfigure[Speaker: no attention]{
    \includegraphics[width=1.15\columnwidth, keepaspectratio]{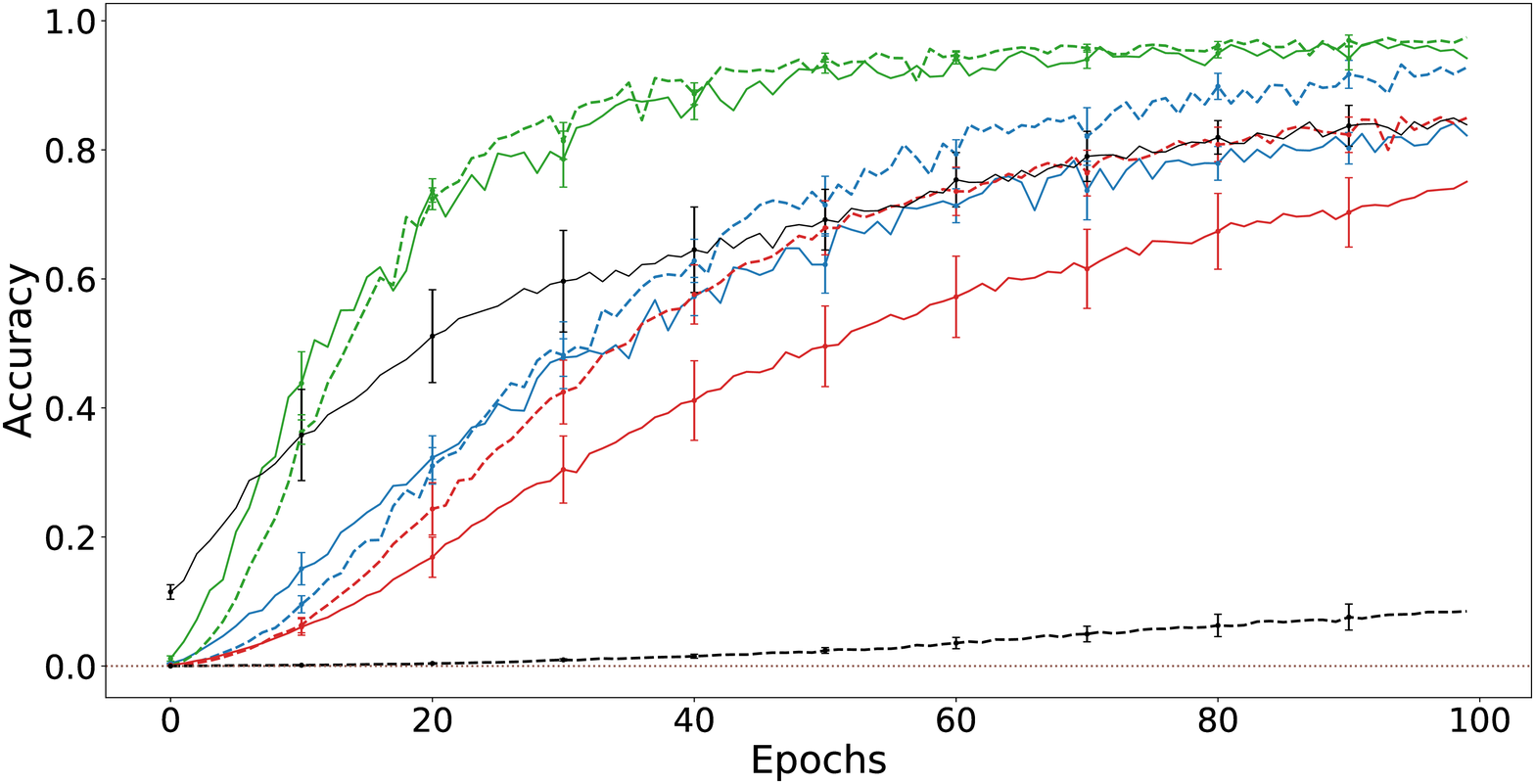}    
    \label{fig:FWO_0_1_T}
}
\vspace{-.8\baselineskip}
\hspace{-2.45\baselineskip}
\subfigure[Listener: no attention]{
    \includegraphics[width=1.15\columnwidth, keepaspectratio]{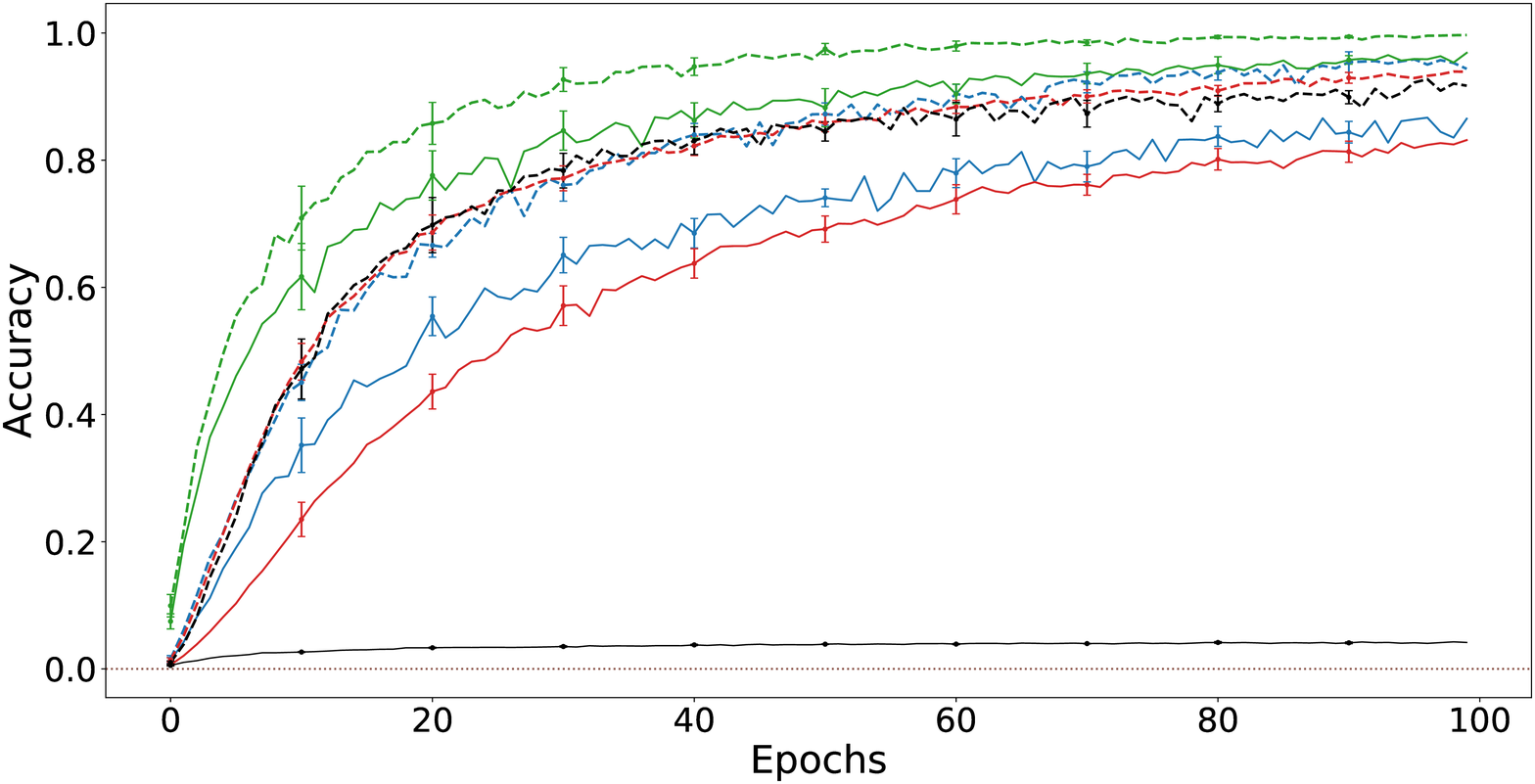} 
    \label{fig:}
}
\subfigure[Speaker: attention  \hspace{2.5\baselineskip}]{
    \hspace{-2.5\baselineskip}	
    \includegraphics[width=1.15\columnwidth, keepaspectratio]{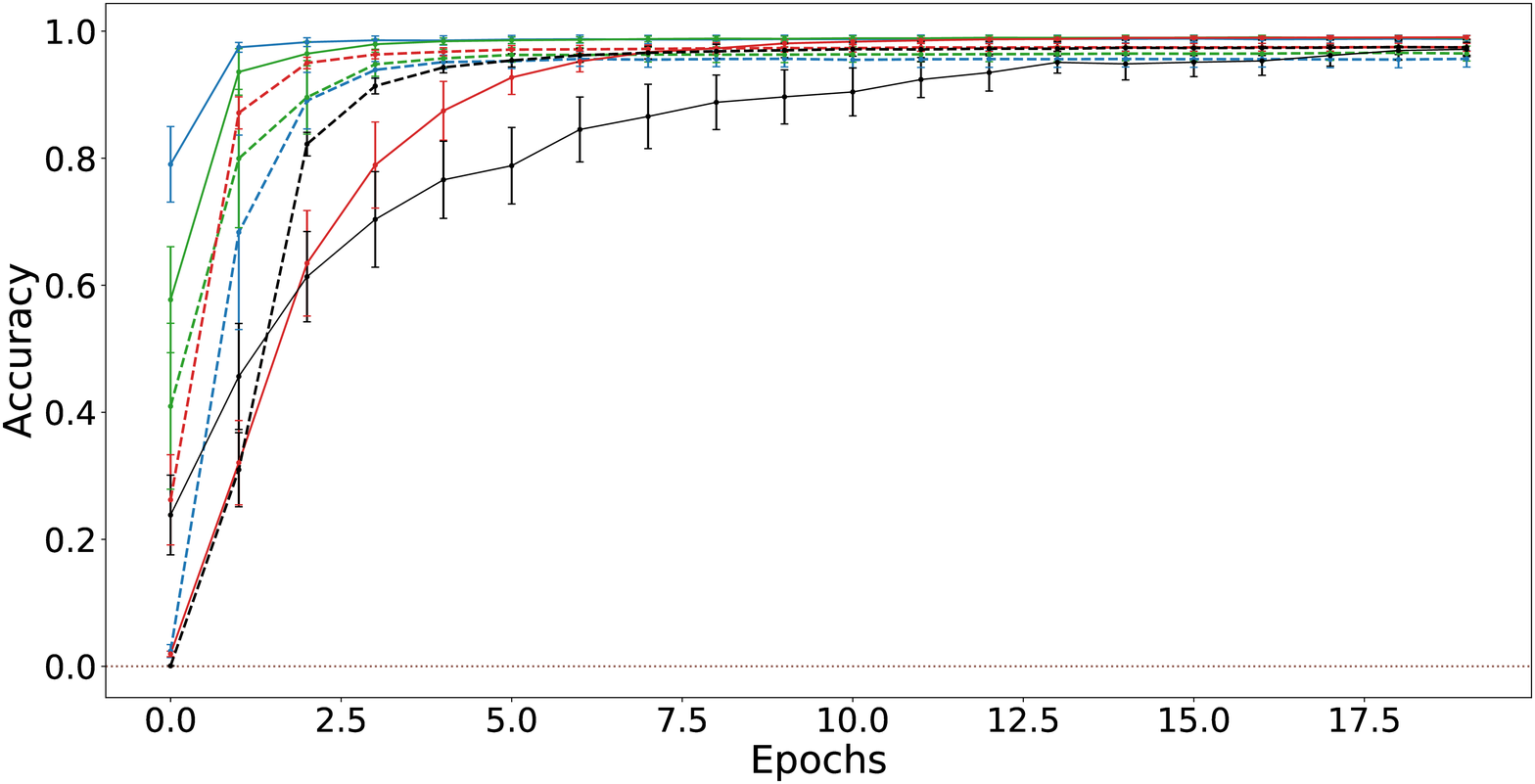}
    \label{fig:FWO_1_1_T}
}
\hspace{-2.5\baselineskip}
\subfigure[Listener: attention]{
   \includegraphics[width=1.15\columnwidth, keepaspectratio]{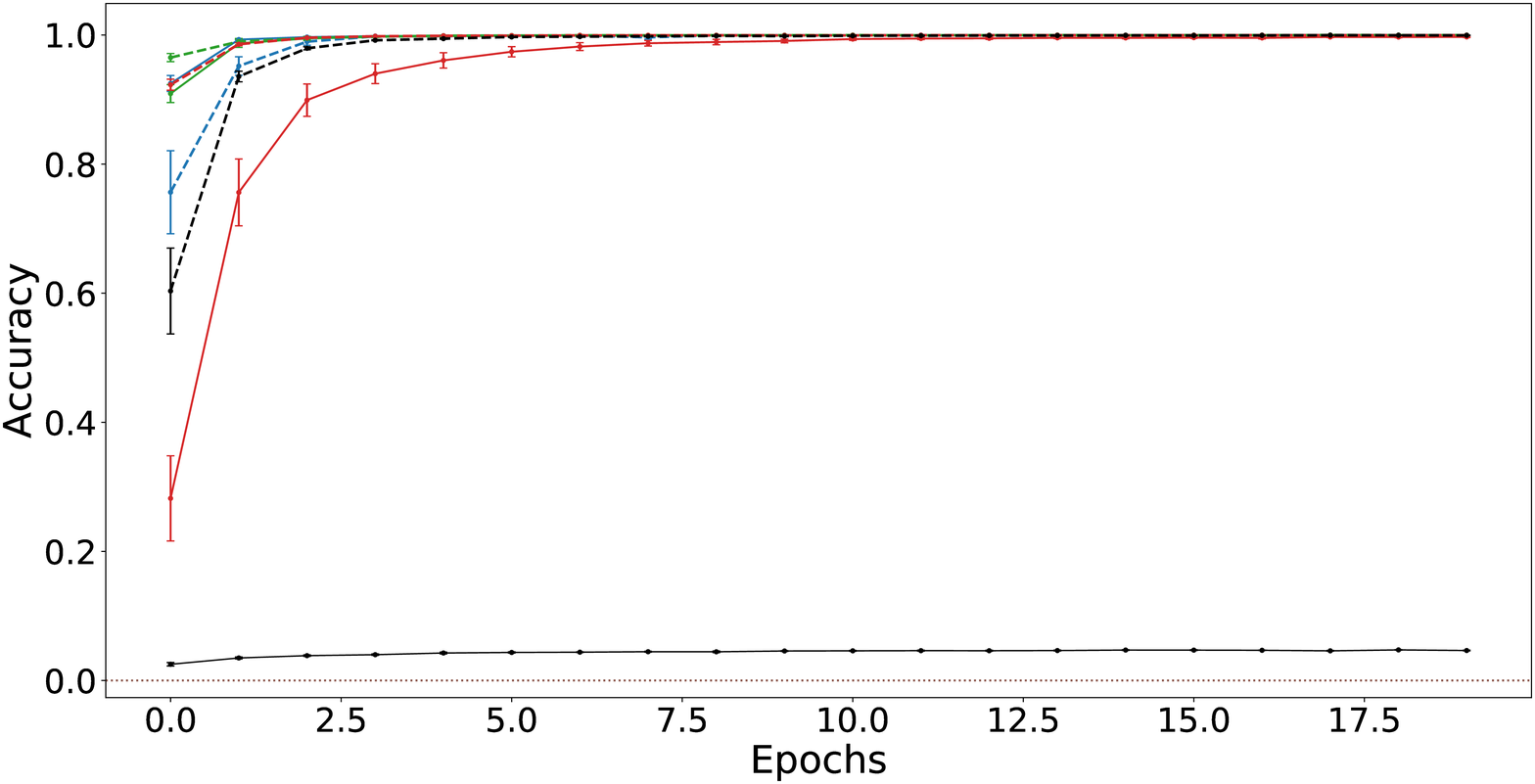}
    \label{fig:FWO_1_1_S}
}
{
    \includegraphics[width=2.1\columnwidth]{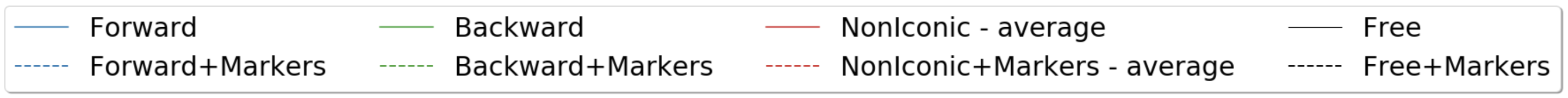}    
}
\vspace{-1.3\baselineskip}
\caption{\textbf{Iconicity / Fixed vs.~free order}: Mean test set
  accuracy in function of training epoch. Error bars represent
  standard deviation over five random seeds.  The NonIconic-average
  curve pools measurements for 3 non-iconic
  languages, each with five runs. Chance  accuracy is
  represented by the horizontal dotted line. The continuous lines represent languages
  without markers, while the dashed lines represent languages with markers.}\label{fig:stats_FWO}
\end{figure*}


\subsection{Iconicity, word order, and markers}
We compare languages with fixed and free order, with and without
markers. Experiments with humans have shown that, as listeners,  children perform better with iconic sentences than non-iconic ones~\cite{DERUITER2018202}. We check whether Seq2Seq networks show similar preferences in terms of learning speed and diachronic persistence.
We compare in particular the forward-iconic order with the backward-iconic language, and three randomly
selected non-iconic languages where the relation between segment and
phrase order is fixed but arbitrary.  Concerning the relation between fixed order and markers, typological studies show a trade-off between these cues. For example, languages with flexible word order  (e.g., Japanese, and Russian) often use case to mark grammatical function, whereas languages with fixed word order (such as English and Mandarin) often lack case marking~\cite[]{Blake:2001, Comrie:1981}. This might be explained by a universal preference for efficient and non-redundant grammatical coding~\cite[]{Fedzechkinaetal2016, qian2012cue, Zipf:1949}. Seq2Seq agents might show similar preferences when tested as Speakers. That is, they might show a learning and preservation preference for either fixed no-marking languages or free marking languages.

\paragraph{Individual learning.} Fig.~\ref{fig:stats_FWO} shows test
accuracy during learning for each language type. The no-attention
agent has a preference for backward-iconic both in speaking and
listening. This is in line with the observation that Seq2Seq machine
translation models work better when the source is presented in reverse
order as it makes the optimization problem easier by introducing shorter-term dependencies~\cite{Sutskever:etal:2014}. The (forward) iconic order is better
than the non-iconic ones in the speaking direction only. The
attention-enhanced model shows much faster convergence to near-perfect
communication, with less room for clear biases to emerge. Still, we
observe some interesting initial preferences. In speaking mode, the
agent learns fastest with the forward iconic language, followed by the
backward one. The non-iconic language without markers is the most
difficult to learn, as expected. On the other hand, in listening mode
we encounter again a preference for backward iconicity.

Only the attention agent in speaking mode shows a trade-off
between order and markers coding, with a preference for markers-free
fixed-order iconic languages over their counterparts with markers, and
for the free-order language with markers over the marker-less
one. Only the non-iconic languages violate the trend: arguably,
though, non-iconic order coding is so sub-optimal that redundant
markers are justified in this case. In listening mode, this agent
shows the expected preference for markers in the free-order case (as
the free-order language without markers is massively ambiguous, with
most utterances mapping to multiple trajectories). However, among the
fixed-order languages, both backward and non-iconic prefer redundant
coding. The agent without attention also displays a preference for
free-order+markers in listening mode (while it has serious difficulties to
learn to speak this language), but no clear avoidance for
redundant coding in either modes. In sum, we confirm a preference for iconic orders. 
Only the attention-enhanced agent in speaking mode displays avoidance of redundant coding.

\paragraph{Iterated learning.} In iterated learning, we might expect
the lineage of agents that starts with less natural non-iconic languages
to either converge to speak more iconic ones, or possibly to drift into low communication accuracy. We
moreover expect redundant coding to fade, with fixed-order+markers
languages to either evolve free order or lose markers. Regarding the free-word order marked language, we expect it
to either converge to a fixed order (possibly iconic) while losing its markers,
as in the historical development from Old English (a language with flexible constituent order and rich case marking) to Modern English (a language with fixed constituent order and a rudimentary case system)~\cite{Traugott1972},
or to remain stable maintaining good communication accuracy.
We focus on the attention agent, as the no-attention one converges too
slowly for multiple-generation experiments. We simulate $10$ generations, repeating each experiment with $5$ different initialization seeds.
For non-iconic orders, we sample the same $3$ languages sampled for individual
learning.

For fixed-order languages, we do not observe any change in accuracy or
behavior in the listener direction (the last-generation child is
perfectly parsing the initial language). However, we observe in speaker mode a (relatively
small) decrease in accuracy across generations, which, importantly,
affects the most natural language (forward iconic without markers) the
least, and the most difficult language (non-iconic without markers)
the most (results are in Supplementary). Again, we observe a (weak) tendency for the attention
agent to yield to the expected natural pressures.

We counted the overall number of markers produced by children in
speaker mode after convergence, for all test trajectories in all
languages with redundant coding. It was always constant,
showing no trend towards losing markers to avoid redundant
coding.  Similarly, there was no tendency, across generations, to start producing multiple utterances in response to the same test trajectory.



\begin{figure}[tb]
   \centering
      \includegraphics[width=\columnwidth]{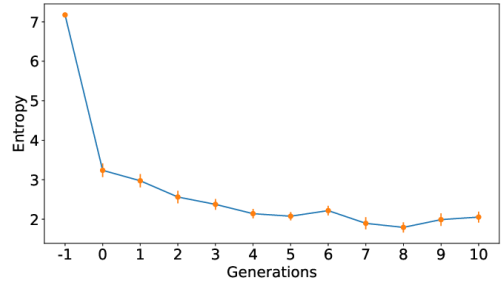}
      \caption{\textbf{Phrase-order entropy} in attention Speaker utterances given test set trajectories, in function of training generation (-1 represents the initial ground-truth distribution). Curve represents mean across $25$ runs, with error bars for standard deviations.}\label{fig:entropy_free}
 \end{figure}

In the evolution of the free-order language with markers, accuracy was
relatively stable in both speaking and listening ($99.82\%$ and $100\%$,
 respectively, for the last-generation agent, averaging across $25$
 runs).\footnote{We run more simulations in this case as we noticed that the final language depends on the initial seed, and hence there is high variance with only $5$ runs. Specifically, we start with $5$ different parents and simulate $10$ generations, repeating each experiment with $5$ different seeds}  
 However, we noticed that across generations, the language becomes more fixed with some preferred orders emerging. 
Fig.~\ref{fig:entropy_free} quantifies this in terms of
the entropy of the observed phrase order probabilities across all test
set trajectories (the lower the entropy, the more skewed the
distribution). There is already a clear decrease for the first agent
with respect to the ground-truth distribution, and the trend continues
across generations.
We analyzed the distribution of Speaker utterances for the longest
($5$-segment) test trajectories in the last generation. We found that, out of $120$ possible phrase orders, no last-generation agent used more than $10$. This is in line with the typological observation that even non-configurational languages favor (at least statistically) certain orders~\cite{Hale:1992,Mithun:1992} and thus an equiprobable distribution of orders, as it is the case in our free word-order+markers language, is unlikely. The ``survivor" orders of the last generation were not necessarily iconic but depended notably on the seed. The absence of clear preference for a specific order could be explained by the fact that attention-enhanced agents, as we saw, can learn any fixed-order language very fast. In this case, the seed of one generation, by randomly skewing the statistics in favor of one order or the other, can significantly impact the preference toward the favored order, that will then spread diachronically throughout the whole iteration.

\subsection{Local vs.\ long-distance} We finally contrast the
long-distance and local languages described in Section
\ref{sec:languages}. In accordance with the linguistic
literature (see Introduction), we predict that the
long-distance language will be harder to learn, and it will tend to
reduce long-distance constructions in diachrony. Although
evidence for distance minimization is typically from production
experiments \cite[e.g.,][]{Futrell:etal:2015}, we expect long-distance
constructions to also be harder in perception, as they cannot be fully
incrementally processed and require keeping material in memory for
longer spans.


\begin{figure*}[ht]
\centering
\hspace{-2.5\baselineskip}
\subfigure[Speaker: attention]
{
    \includegraphics[width=1.15\columnwidth, keepaspectratio]{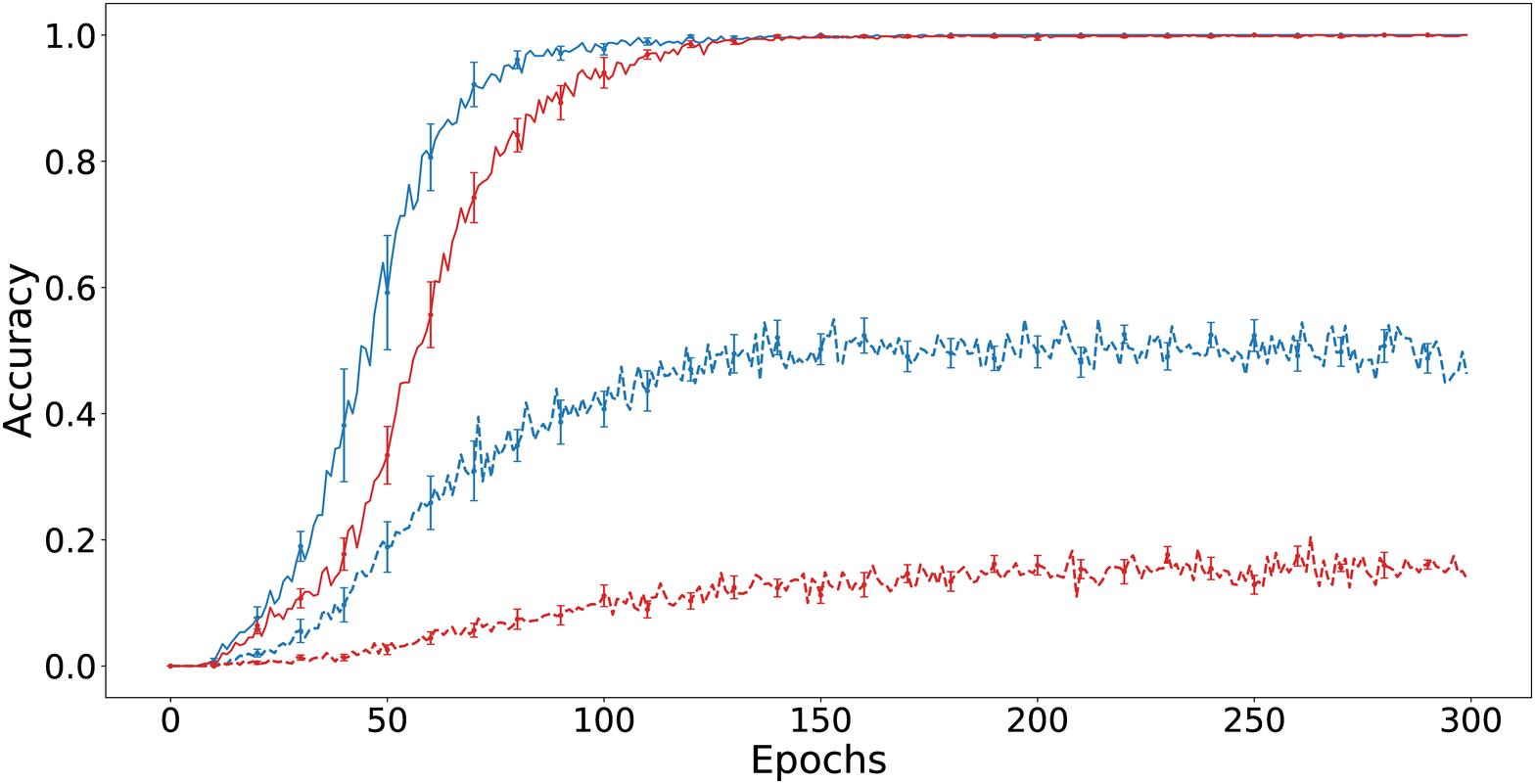}    
    \label{fig:DM_1_1_teacher}
}
\hspace{-2.45\baselineskip}
\subfigure[Listener: attention]
{
    \includegraphics[width=1.15\columnwidth, keepaspectratio]{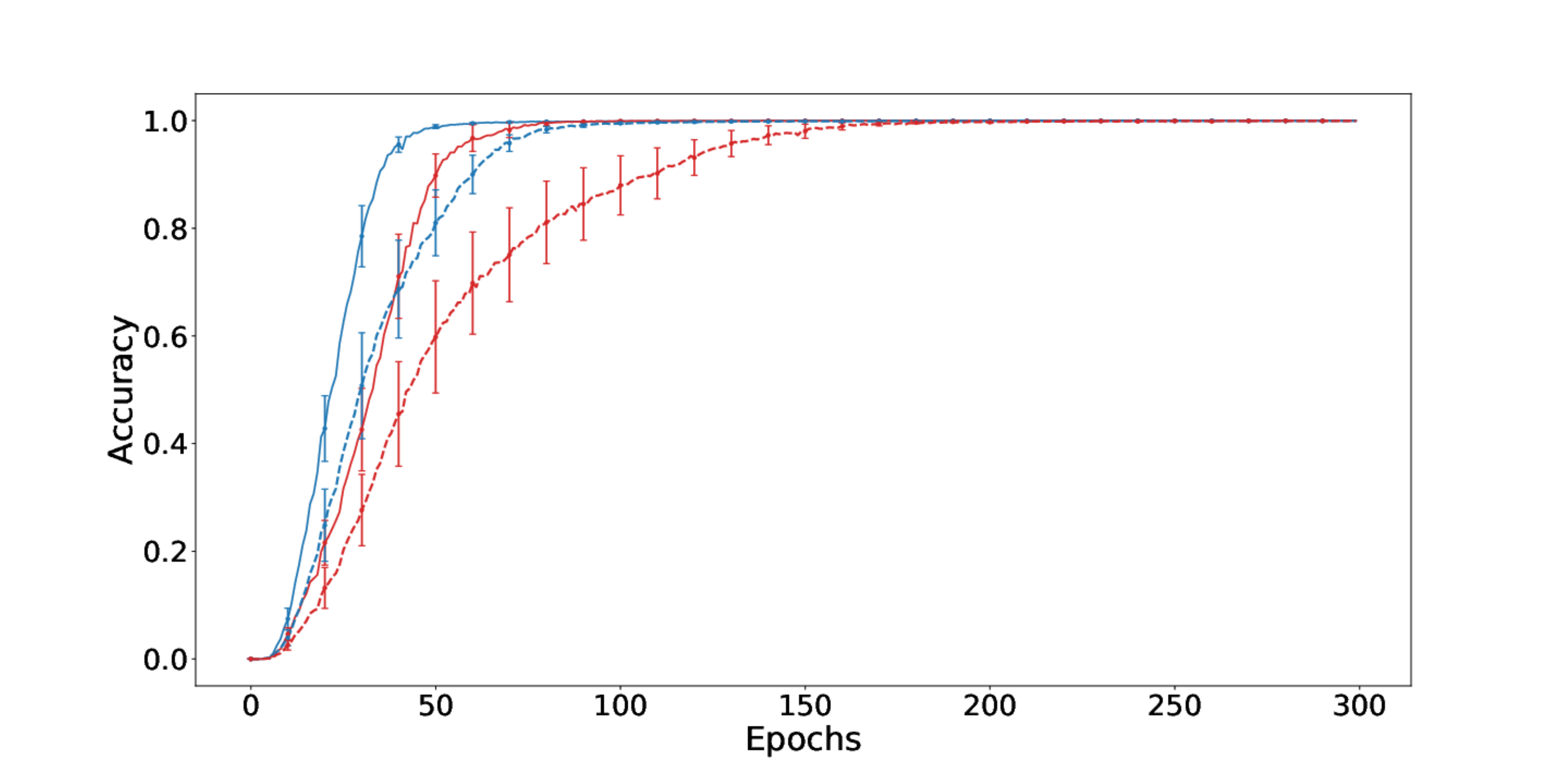}       
    \label{fig:DM_1_1_student}
}
{
    \includegraphics[width=1.5\columnwidth]{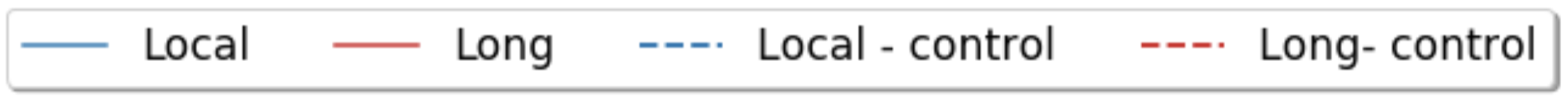}    
}
\vspace{-.3\baselineskip}
\caption{\textbf{Long vs.\ local distance}: Mean test set accuracy as a function of training epoch. The error bars correspond to the standard deviation, calculated over five random seeds.}\label{fig:stats_DM}
\end{figure*}

\paragraph{Individual learning.} As the long-distance language includes all utterances from the local language, it might be trivially
harder to learn. To account for this, we construct a set of
\emph{control} languages by randomly sampling, for each trajectory,
the same number of possible utterances for the local and long-distance
controls. We report
averaged results for 3 such languages of both kinds. Details on their
construction are in Supplementary.

Fig.~\ref{fig:stats_DM} shows test set accuracy across $300$ training epochs for the attention model. The results, for
speaking and listening, confirm the preference for the local
language. The control languages are harder to learn, as they impose an
arbitrary constraint on free word order, but they display the
preference for the local language even more clearly. Overall, we see a
tendency for listening to be easier than speaking, but this cuts
across the local/long-distance division, and it seems to be a more
general consequence of free-order languages with markers being easier in parsing
than production (cf.~the no-attention agent results in Fig.~\ref{fig:stats_FWO}). Results without attention
(not shown) are comparable in general, although the listener/speaker
asymmetry is sharper, with no difference in difficulty among the
4 languages when listening.

\begin{figure}
\centering
{
	\includegraphics[width=\columnwidth]{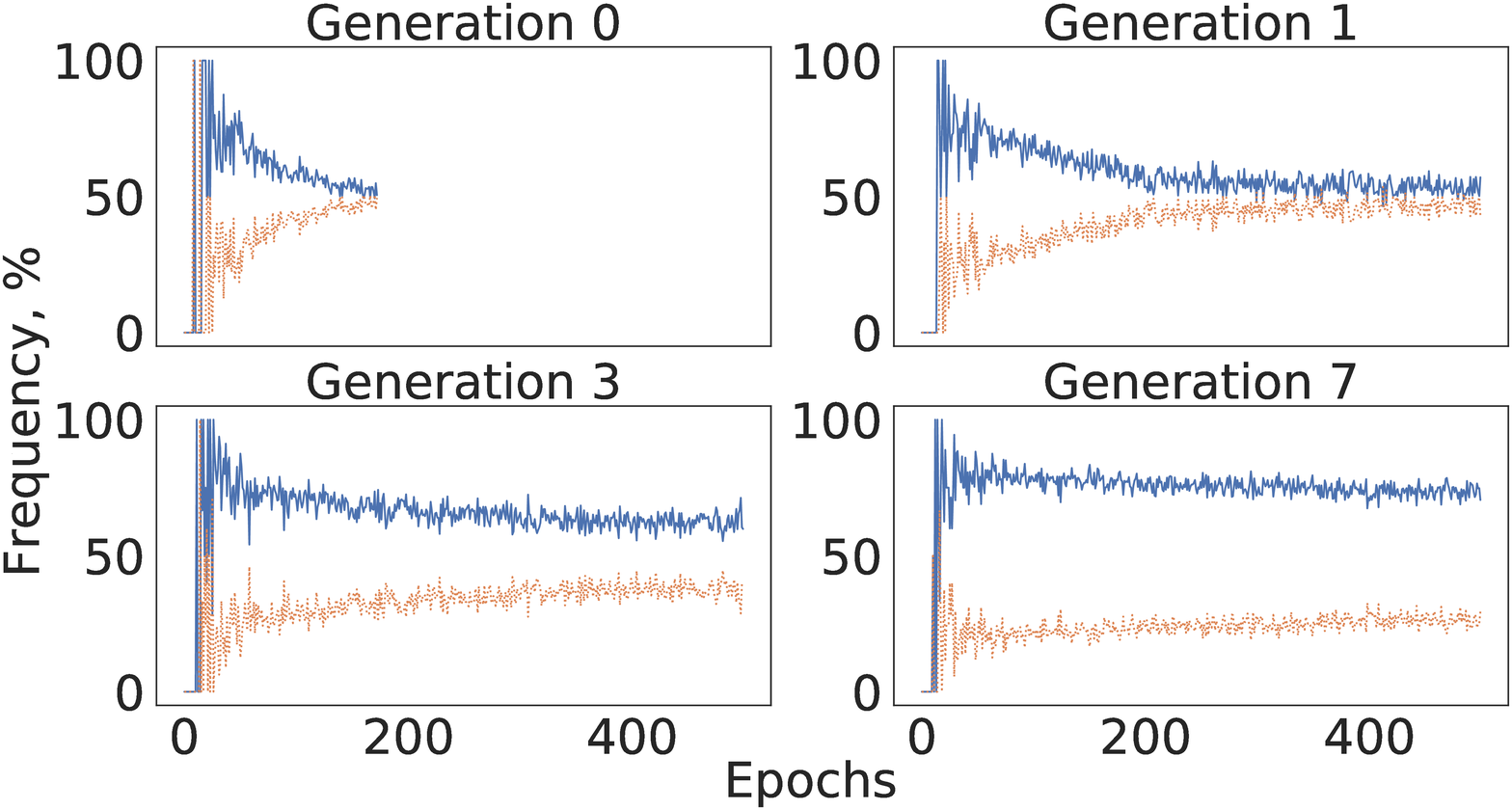}

}

{
    \includegraphics[width=\columnwidth]{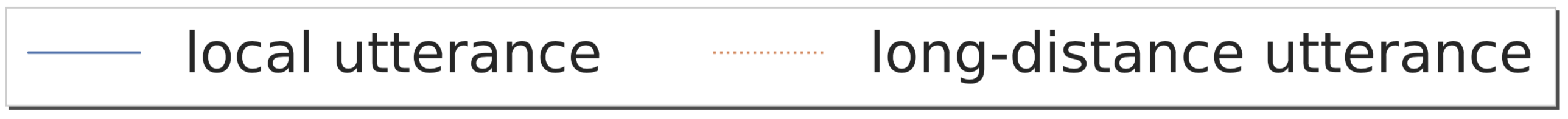}    
}

\vspace{-1\baselineskip}
\caption{Frequency of the local and long-distance utterances produced by the attention Speaker in function of training epoch. The input trajectories are taken from the test set. Test set accuracies for the four generations shown: $99.99\%$, $87.62\%$, $84.54\%$, $79.38\%$. At Generation 0, less epochs were run due to early stopping. \vspace{-2\baselineskip}}\label{fig:DM_iterated}
\end{figure}

\paragraph{Iterated learning.} We study multiple-generation
transmission of the long-distance language with the attention
agent. To deal with the problem of skewed relative frequency of
long-distance and entirely local utterances, the Speaker direction is
trained by ensuring that the output utterance set $\{\vu\}$ for each
input trajectory $\vt$ contains the same number of long-distance and
local constructions. This is achieved by sub-sampling $n=48$
long-distance utterances to match the number of possible local
constructions. Fig.~\ref{fig:DM_iterated} shows the relative frequency
across generations of local and long-distance utterances produced by
the agent as a Speaker in function of training (one
representative seed of 5). As predicted, a clear
preference for local constructions emerges, confirming the presence of
a distance minimization bias in Seq2Seq models.

\section{Discussion}
\label{sec:conclusion}

We studied whether word-order constraints widely attested in natural
languages affect learning and diachronic transmission in Seq2Seq
agents. We found that some trends follow natural patterns, such as the
tendency to limit word order to few configurations, and long-distance
dependency minimization. In other ways, our agents depart from typical
human language patterns. For example, they exhibit a preference for a
backward order, and there are only weak signs of a trade-off
between different ways to encode constituent roles, with redundant
solutions often being preferred.

The research direction we introduced might lead to a better
understanding of the biases that affect the linguistic
behaviour of LSTMs and similar models. This could 
help current efforts towards the development of artificial
agents that communicate to solve a task, with the ultimate goal of
developing AIs that can talk with humans. It has been observed
that the communication protocol emerging in such simulations is very
different from human language
\cite[e.g.,][]{Kottur:etal:2017,Lewis:etal:2017,Bouchacourt:Baroni:2018}.
A better understanding of what are the ``innate'' biases of standard models in 
highly controlled setups, such as the one studied here, should complement large-scale 
simulations, as part of the effort to develop new methods to encourage the emergence 
of more human-like language. For example, our results suggest that current neural networks, 
as they are not subject to human-like least-effort constraints, might not display the same trend towards efficient communication 
that we encounter in natural languages. How to incorporate ``effort''-based pressures in neural networks is an exciting direction for future work.

\section{Acknowledgments}
We would like to thank Roger Levy, Diane Bouchacourt, Alex Cristea, Kristina Gulordava and Armand Joulin for their very helpful feedback.
\bibliography{marco_uniform,other}
\bibliographystyle{acl_natbib}

\section{Supplementary Material}

\subsection{Deriving the training loss for the Speaker role}
\label{sec:supp:Loss}
In the Speaker role, each input trajectory $\vt_j$ maps onto a set of utterances $\{\vu\}_j$. We want to train an agent such that,
given $\vt_j$, it generates all the corresponding utterances $\{\vu\}_j$ uniformly. To do that, we follow the ``Na\"ive'' approach from \newcite{Jin2003}.

Given an input 
$\vt$, the Seq2Seq model defines a distribution over the output sequences, $p_{\theta}(\vu | \vt_j)$.
The KL-divergence~\cite{Kullback1997} $\mathcal{D}(P || p_{\theta})$ between the uniform distribution $P(\vu | \vt_j)$ over the target utterances $\{\vu\}_j$ and the output distribution of the agent, $p_{\theta}(\vu | \vt_j)$, is:
\begin{equation}
\begin{split}
    \mathcal{D}(P || p_{\theta}) = & \mathbb{E}_{\vu \sim P(\vu| \vt_j)} \left[ \log \frac{P(\vu| \vt_j)}{p_{\theta}(\vu| \vt_j)} \right] \\
        = & E - \mathbb{E}_{\vu \sim P(\vu| \vt_j)} \log p_{\theta}(\vu| \vt_j)
\end{split}
\end{equation}
with $E$ independent from $\theta$. Hence, finding $\theta$ that  minimizes $\mathcal{D}(P || p_{\theta})$ is equivalent to minimization of $\mathcal{L}'$:
\begin{equation}
    \label{eq:supp:expectation}
\mathcal{L}'(\vt_j) = -\mathbb{E}_{\vu \sim P(\vu| \vt_j)} \log p_{\theta}(\vu| \vt_j)
\end{equation}

Next, assuming that the target set of utterances $\{\vu_j\}$ has $n_j$ elements,
\begin{equation}
\label{eq:supp:expand_div}
\mathbb{E}_{\vu \sim P(\vu| \vt_j)} \log p_{\theta}(\vu| \vt_j) = \frac{1}{n_j} \sum_{\vu \in \{\vu\}_j } \log p_{\theta}(\vu | \vt_j)
\end{equation}
We expand $p_{\theta}(\vu| \vt_j)$ by iterating words $u_k$ in $\vu$, as in Section 3.2 of the main text:
\begin{equation}
\label{eq:supp:expand_prob}
\log p_{\theta}(\vu | \vt_j) = \sum_{k=1}^{|\vu|} \log p_{\theta}(u_{k} | u_{k-1}, \vh_{j, k})
\end{equation}
By combining Eq.~(\ref{eq:supp:expand_div}) and Eq.~(\ref{eq:supp:expand_prob}), we obtain:

\begin{equation}
\label{eq:supp:loss}
    \mathcal{L}'(\vt_j) = -\frac{1}{n_j} \sum_{\vu \in \{\vu\}_j } \sum_{k=1}^{|\vu|} \log p_{\theta}(u_{k} | u_{k-1}, \vh_{j, k})
\end{equation}
After aggregation over all trajectories in the dataset, we obtain the full loss that coincides with Eq.~(4) in the main text:
\begin{equation}
    \begin{split}
\label{eq:supp:loss_full}
        \sum_{j} & \mathcal{L}'(\vt_j) = \\
        & -\sum_{j}\frac{1}{n_j}  \sum_{\vu \in \{\vu\}_j } \sum_{k=1}^{|\vu|} \log p_{\theta}(u_{k} | u_{k-1}, \vh_{j, k})
    \end{split}
\end{equation}
This concludes our derivation of the loss used for the Speaker role. 
Finally, we note that this derivation provides grounding for the sub-sampling we use during the training, as it corresponds to getting a Monte-Carlo estimate of the expectation in Eq.~(\ref{eq:supp:expectation}) over $n$ samples, instead of the full support of the distribution.

\subsection{Examples of trajectories and utterances}
In Table~\ref{supp:table:examples}, we exemplify how trajectories with two, three, and five segments are represented by utterances in free- and fixed-order languages with and without markers. Note how the free-order language without markers is extremely ambiguous, as the utterances do not encode the execution order of the corresponding trajectories.

Tables \ref{supp:table:examples_local} and
\ref{supp:table:examples_long} give examples of how trajectories are
represented by utterances in the local and long-distance languages, as
well as in example controls. The control languages are constructed to
enable a fairer comparison between the local and long-distance
setups. The full long-distance language has more possible utterances
per trajectory than the local one (the latter is a subset of the
former). Their controls, however, have the same number of
utterances. Practically, to construct one local control language, we
sample $24$ distinct utterance templates (that is, phrase orders) out
of $48$ from the full language.  We use $3$ different local control
languages by sampling a different subset each time. Table
\ref{supp:table:examples_local} exemplifies one of these control
languages.  To construct one long-distance control language, we also
sample $24$ distinct utterance templates from the full long-distance
language (out of $144$ possible utterances).  The latter sampling
maintains the proportion of local and long-distance constructions of
the full long-distance language ($1/3$ vs.~$2/3$). Again, we sample
$3$ different long-distance controls. One of them is exemplified in
Table \ref{supp:table:examples_long}.


\begin{table*}[tb]
    \caption{Example utterances associated to trajectories of different lengths in the fixed- and free-order languages we consider.} \label{supp:table:examples}
    \centering
\begin{adjustbox}{width=1\textwidth}
\begin{tabular}{lccccccccccccr}
\toprule
Trajectory (two segments): & LEFT DOWN DOWN \\
\midrule
With markers & \textbf{Forward-iconic}              & \textbf{Reverse-iconic} & \textbf{Non-iconic 1} \\ 
&  first left 1 second down 2 & second down 2 first left 1 & first left 1 second down 2 \\
& \textbf{Non-iconic 2} & \textbf{Non-iconic 3} & \textbf{Free-order} \\
&  second down 2 first left 1 &  first left 1 second down 2 & first left 1 second down 2 \\
&   &  & second down 2 first left 1 \\
\midrule
Without markers & \textbf{Forward-iconic}              & \textbf{Reverse-iconic} & \textbf{Non-iconic 1} \\ 
&  left 1 down 2 & down 2 left 1 & left 1 down 2 \\
& & & \textbf{Free-order} \\
& \textbf{Non-iconic 2} & \textbf{Non-iconic 2} & left 1 down 2 \\
&  down 2 left 1  & left 1 down 2  & down 2 left 1 \\
\midrule
Trajectory (three segments): & LEFT LEFT LEFT DOWN DOWN UP UP UP \\
\midrule
With markers &   & & \textbf{Non-iconic 1} \\ 
& \textbf{Forward-iconic} & \textbf{Reverse-iconic} & first left 3 third up 3 second down 2 \\
& first left 3 second down 2 third up 3 & third up 3 second down 2 first left 3  & \textbf{Free-order} \\
& & & first left 3 second down 2 third up 3 \\
& & & first left 3 third up 3 second down 2  \\
& \textbf{Non-iconic 2} & \textbf{Non-iconic 3} & second down 2 third up 3 first left 3   \\
& second down 2 third up 3 first left 3 & first left 3 second down 2 third up 3 & second down 2 first left 3 third up 3 \\
& & & third up 3 first left 3 second down 2  \\
& & & third up 3 second down 2 first left 3  \\
\midrule
Without markers &  & & \textbf{Non-iconic 1} \\ 
& \textbf{Forward-iconic} & \textbf{Reverse-iconic} & left 3 up 3 down 2 \\
& left 3 down 2 up 3 & up 3 down 2 left 3  & \textbf{Free-order} \\
& & & left 3 down 2 up 3 \\
& & & left 3 up 3 down 2  \\
& \textbf{Non-iconic 2} & \textbf{Non-iconic 3} & down 2 up 3 left 3   \\
& down 2 up 3 left 3 & left 3 down 2 up 3 & down 2 left 3 up 3 \\
& & & up 3 left 3 down 2  \\
& & & up 3 down 2 left 3  \\
\midrule
Trajectory (five segments): & DOWN RIGHT RIGHT UP UP UP RIGHT LEFT LEFT \\
\midrule
With markers & \textbf{Forward-iconic}              & \textbf{Reverse-iconic} & \textbf{Non-iconic 1} \\ 
& first down 1 second right 2 third up 3 fourth right 1 fifth left 2 & fifth left 2 fourth right 1 third up 3 second right 2 first down 1 & first down 1 fourth right 1 third up 3 second right 2 fifth left 2 \\
& &  & \textbf{Free-order} \\
& & & first down 1 second right 2 third up 3 fourth right 1 fifth left 2 \\
& \textbf{Non-iconic 2} & \textbf{Non-iconic 3} & first down 1 second right 2 third up 3 fifth left 2 fourth right 1 \\
& second right 2 third up 3 fifth left 2 fourth right 1 first down 1 & fourth right 1 first down 1 second right 2 fifth left 2 third up 3 & \ldots \\
& & & fifth left 2 fourth right 1 third up 3 second right 2 first down 1 \\
\midrule Without markers & &  & \textbf{Non-iconic 1} \\ 
& \textbf{Forward-iconic}              & \textbf{Reverse-iconic} & down 1 right 1 up 3 right 2 left 2 \\
& down 1 right 2 up 3 right 1 left 2 & left 2 right 1 up 3 right 2 down 1 & \textbf{Free-order} \\
& & & down 1 right 2 up 3 right 1 left 2 \\
& \textbf{Non-iconic 2} & \textbf{Non-iconic 3} & down 1 right 2 up 3 left 2 right 1 \\
& right 2 up 3 left 2 right 1 down 1 & right 1 down 1 right 2 left 2 up 3 & \ldots \\
& & & left 2 right 1 up 3 right 2 down 1 \\
\bottomrule
\end{tabular}
\end{adjustbox}
\end{table*}

\begin{table*}[tb]
    \caption{Example utterances associated to one trajectory by the local language and one of its controls.}\label{supp:table:examples_local}
    \centering
    \begin{adjustbox}{width=1\textwidth}
        \begin{tabular}{p{9cm} p{9cm}p{9cm}}
\toprule
            \multicolumn{3}{l}{Trajectory (three segments): DOWN DOWN DOWN LEFT LEFT LEFT UP} \\
\midrule
            \multicolumn{3}{c}{\textbf{Local}} \\
    first down first 3 second left second 3 third up third 1 & first down first 3 second left second 3 third 1 third up & first down first 3 second 3 second left third up third 1 \\
    first down first 3 second 3 second left third 1 third up & first down first 3 third up third 1 second left second 3 & first down first 3 third up third 1 second 3 second left \\
            ~~~~~~~~~~~~~~~~~~~~~~~~~~~~~~~~~~~~~~~~~~~ \ldots & ~~~~~~~~~~~~~~~~~~~~~~~~~~~~~~~~~~~~~~~~~~~ \ldots &  ~~~~~~~~~~~~~~~~~~~~~~~~~~~~~~~~~~~~~~~~~~~ \ldots \\
    third 1 third up first 3 first down second left second 3 & third 1 third up first 3 first down second 3 second left & third 1 third up second left second 3 first down first 3 \\
    third 1 third up second left second 3 first 3 first down & third 1 third up second 3 second left first down first 3 & third 1 third up second 3 second left first 3 first down \\
\midrule
    \multicolumn{3}{c}{\textbf{Local control} (one of three)} \\
    first down first 3 second left second 3 third up third 1 & first down first 3 third up third 1 second left second 3 & second 3 second left third up third 1 first down first 3 \\
    second 3 second left third 1 third up first down first 3 & third up third 1 second left second 3 first down first 3 & second left second 3 first 3 first down third 1 third up \\
    second left second 3 first down first 3 third up third 1 & first 3 first down third up third 1 second 3 second left & third up third 1 first 3 first down second 3 second left \\
    third up third 1 first down first 3 second left second 3 & third 1 third up second left second 3 first 3 first down & third 1 third up second 3 second left first 3 first down \\
    second 3 second left first 3 first down third up third 1 & second 3 second left third up third 1 first 3 first down & first 3 first down third up third 1 second left second 3 \\
    second 3 second left first down first 3 third 1 third up & third 1 third up first 3 first down second 3 second left & first down first 3 third 1 third up second left second 3 \\
    third up third 1 second 3 second left first 3 first down & third up third 1 first down first 3 second 3 second left & third up third 1 second left second 3 first 3 first down \\
    first 3 first down second 3 second left third up third 1 & first 3 first down second left second 3 third up third 1 & second left second 3 first down first 3 third 1 third up \\
\bottomrule
\end{tabular}
\end{adjustbox}
\end{table*}

\begin{table*}[tb]
    \caption{Example utterances associated to one trajectory by the long-distance language and one of its controls.}\label{supp:table:examples_long}
    \centering
    \begin{adjustbox}{width=1\textwidth}
\begin{tabular}{p{9cm} p{9cm}p{9cm}}
\toprule
     \multicolumn{3}{l}{Trajectory (three segments): DOWN DOWN DOWN LEFT LEFT LEFT UP} \\
\midrule
     \multicolumn{3}{c}{\textbf{Long-distance}} \\
\midrule
     \multicolumn{3}{c}{\textit{local utterances}} \\
first down first 3 second left second 3 third up third 1 & first down first 3 second left second 3 third 1 third up & first down first 3 second 3 second left third up third 1 \\
first down first 3 second 3 second left third 1 third up & first down first 3 third up third 1 second left second 3 & first down first 3 third up third 1 second 3 second left \\
            ~~~~~~~~~~~~~~~~~~~~~~~~~~~~~~~~~~~~~~~~~~~ \ldots & ~~~~~~~~~~~~~~~~~~~~~~~~~~~~~~~~~~~~~~~~~~~ \ldots &  ~~~~~~~~~~~~~~~~~~~~~~~~~~~~~~~~~~~~~~~~~~~ \ldots \\
third 1 third up first 3 first down second left second 3 & third 1 third up first 3 first down second 3 second left & third 1 third up second left second 3 first down first 3 \\
third 1 third up second left second 3 first 3 first down & third 1 third up second 3 second left first down first 3 & third 1 third up second 3 second left first 3 first down \\
\midrule
     \multicolumn{3}{c}{\textit{long-distance utterances}} \\
first down first 3 second left third up third 1 second 3 & first down first 3 second left third 1 third up second 3 & first down first 3 second 3 third up third 1 second left \\
first down first 3 second 3 third 1 third up second left & first down first 3 third up second left second 3 third 1 & first down first 3 third up second 3 second left third 1 \\
            ~~~~~~~~~~~~~~~~~~~~~~~~~~~~~~~~~~~~~~~~~~~ \ldots & ~~~~~~~~~~~~~~~~~~~~~~~~~~~~~~~~~~~~~~~~~~~ \ldots &  ~~~~~~~~~~~~~~~~~~~~~~~~~~~~~~~~~~~~~~~~~~~ \ldots \\
third 1 third up first 3 second left second 3 first down & third 1 third up first 3 second 3 second left first down & third 1 third up second left first down first 3 second 3 \\
third 1 third up second left first 3 first down second 3 & third 1 third up second 3 first down first 3 second left & third 1 third up second 3 first 3 first down second left \\
\midrule
    \multicolumn{3}{c}{\textbf{Long-distance control} (one of three)} \\
\midrule
     \multicolumn{3}{c}{\textit{local utterances}} \\
first down first 2 second left second 3 third up third 1 & second 3 second left first 2 first down third up third 1 & third up third 1 first down first 2 second left second 3 \\
third 1 third up first down first 2 second 3 second left & second left second 3 first down first 2 third 1 third up & second 3 second left third 1 third up first down first 2 \\
second 3 second left third up third 1 first 2 first down & second left second 3 first down first 2 third up third 1 & \\
\midrule
     \multicolumn{3}{c}{\textit{long-distance utterances}} \\
first 2 first down third up second 3 second left third 1 & third 1 second 3 second left third up first down first 2 & first 2 third up third 1 first down second left second 3 \\
first 2 second 3 second left first down third 1 third up & second 3 third up third 1 second left first 2 first down & second 3 first down first 2 second left third 1 third up \\
second 3 second left first down third 1 third up first 2 & third 1 second left second 3 third up first 2 first down & second left second 3 third up first 2 first down third 1 \\
third 1 third up first 2 second 3 second left first down & first down second left second 3 first 2 third 1 third up & third 1 first down first 2 third up second left second 3 \\
third up third 1 second left first 2 first down second 3 & third up third 1 second 3 first down first 2 second left & third 1 third up second left first 2 first down second 3 \\
    &    first 2 first down third 1 second 3 second left third up & \\

\bottomrule 
\end{tabular}
\end{adjustbox}
\end{table*}

\subsection{Iterated Learning of fixed word order languages}
\begin{figure}[tb]
\centering
   \includegraphics[width=\columnwidth]{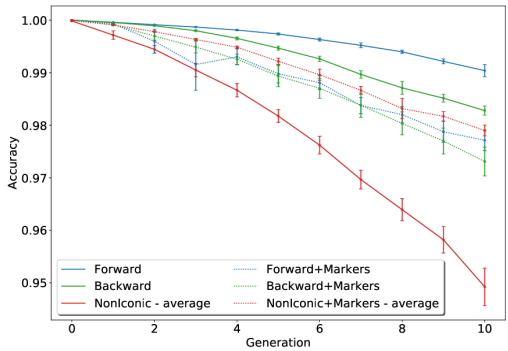}
    \vspace{-2\baselineskip}
    \caption{\textbf{Iterated learning with fixed-order languages.} Mean test set attention Speaker accuracy at the end of training over $10$ generations. Error bars represent standard deviation over $5$ random seeds. The NonIconic-average curve pools measurements for $3$ non-iconic languages, each with $5$ runs. \vspace{-1\baselineskip}}\label{fig:fixed_decay}
\end{figure}

In this section, we use the iterated learning paradigm to analyze Seq2Seq networks biases toward iconic languages. We expect agents with less natural non-iconic languages to either converge to more iconic ones or diverge with low communication accuracy. We simulate $10$ generations repeating the process with $5$ different initialization seed and report the average of communication accuracy of each generation in Fig.~\ref{fig:fixed_decay}. We observe in speaker mode a (relatively small) decrease in accuracy across generations, which, importantly,
affects the most natural language (forward iconic without markers) the
least, and the most difficult language (non-iconic without markers) the most. Thus, we observe a (weak) tendency for the attention
agent to yield to the expected natural pressures in terms of iconic order.

\end{document}